\documentclass[10pt,twocolumn,letterpaper]{article}

\usepackage{cvpr}
\usepackage{times}
\usepackage{epsfig}
\usepackage{graphicx}
\usepackage{amsmath}
\usepackage{amssymb}

\usepackage{capt-of}
\usepackage{adjustbox}
\usepackage{booktabs}
\usepackage{sidecap}
\usepackage{amsfonts}
\usepackage{multirow}
\usepackage[table,xcdraw]{xcolor}
\usepackage{authblk}

\usepackage[breaklinks=true,bookmarks=false]{hyperref}

\cvprfinalcopy 

\def\cvprPaperID{****} 

\begin{document}

\title{Dynamic Adaptation on Non-Stationary Visual Domains}


 \author[1,2]{Sindi Shkodrani      
      \thanks{\tt sindi.shkodrani@tomtom.com}}
\author[2]{Michael Hofmann \thanks{\tt michael.hofmann@tomtom.com}}
\author[1]{Efstratios Gavves \thanks{\tt e.gavves@uva.nl}}

\affil[1]{University of Amsterdam}
\affil[2]{TomTom, Amsterdam, The Netherlands}


\maketitle

\begin{abstract}

Domain adaptation aims to learn models on a supervised source domain that perform well on an unsupervised target. Prior work has examined domain adaptation in the context of stationary domain shifts, i.e.~static data sets. However, with large-scale or dynamic data sources, data from a defined domain is not usually available all at once. For instance, in a streaming data scenario, dataset statistics effectively become a function of time. We introduce a framework for adaptation over non-stationary distribution shifts applicable to large-scale and streaming data scenarios. The model is adapted sequentially over incoming unsupervised streaming data batches. This enables improvements over several batches without the need for any additionally annotated data. To demonstrate the effectiveness of our proposed framework, we modify associative domain adaptation to work well on source and target data batches with unequal class distributions. We apply our method to several adaptation benchmark datasets for classification and show improved classifier accuracy not only for the currently adapted batch, but also when applied on future stream batches. Furthermore, we show the applicability of our associative learning modifications to semantic segmentation, where we achieve competitive results.

\end{abstract}
\section{Introduction}
\label{sec:intro}

Domain adaptation aims to adapt classifiers trained on source domains to novel unlabeled target domains, where a domain shift, namely a difference in distribution statistics, is expected~\cite{survey_da,  TommasiPCT15}. Typically, the domain shift is considered within the context of ``closed'', static domains, implicitly assuming datasets available in their entirety at adaptation time ~\cite{survey_wang, survey_da}.
However, in realistic applications data collection is not static nor closed but ``open'', giving rise to non-stationary domain shifts \cite{active_learning}.

\begin{figure}[t]
 \includegraphics[width=.99\linewidth]{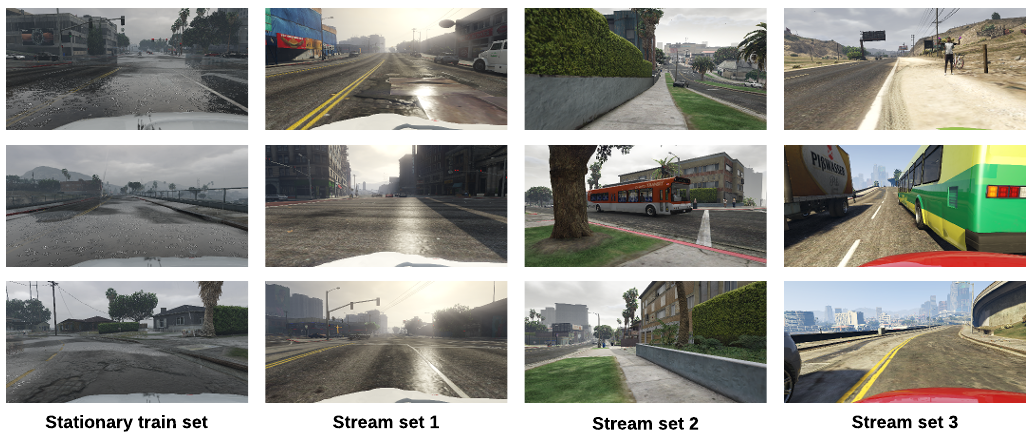}
  \caption{Distribution shift across stream batches in GTA5}
  \label{fig:distrib_shift}
\end{figure}

Consider e.g~ social media feeds, or urban imagery taken from inside a car. These images often arrive in ``bundles'' with different distribution statistics, due to, for instance, being collected in different cities or with different weather conditions (see Figure \ref{fig:distrib_shift}). If we were to consider these bundles as isolated domains, we wouldn't be exploiting the available unlabeled data entirely. In addition, if the distribution changes gradually over time in a streaming-like fashion, being able to adapt over bundles sequentially may benefit real-time predictions on future incoming data bundles. In streaming data this distribution shift over time is called concept drift, and the incoming stream is usually too large to be held in memory, therefore it is processed in data bundles which are later discarded.  

As an adaptation method, we look into associative learning proposed by Haeusser \emph{et al.}~\cite{learning_by_assoc,assoc_da}, which uses association of embeddings in latent space and has been shown to work well for domain adaptation and semi-supervised learning. However, associative domain adaptation makes the implicit assumption that the class probability distributions between the source and the target domains are similar at adaptation time. 
This assumption cannot be guaranteed to hold when the target dataset is not well known in advance, such as in ``open'' datasets the class probability statistics may change dynamically or in tasks where class statistics across domains may vary a lot. An example of such a task is semantic segmentation. 
To this end, the associations between source and target embeddings need to be performed while taking into account the non-stationary changes of the class probability statistics.

\begin{figure}[t]
\includegraphics[width=.99\linewidth]{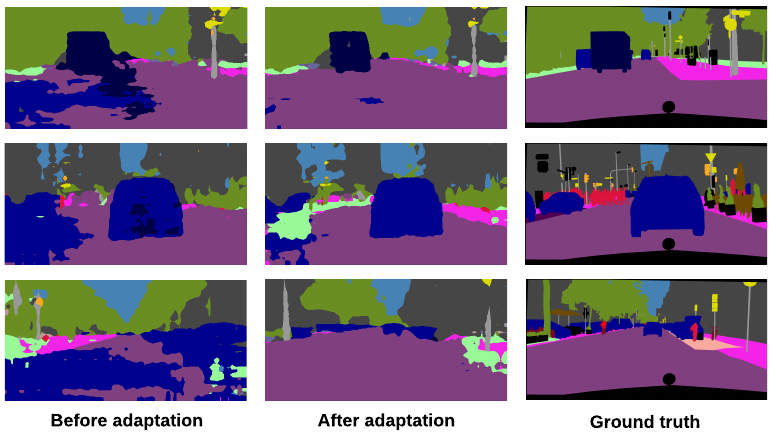}
\caption{Adaptation results for semantic segmentation on Cityscapes}
\label{fig:seg_adapt}
\end{figure}

This work makes three contributions.
First, we argue that domain adaptation is important beyond static domain datasets, including continuously collected datasets whose statistics are non-stationary. For dynamic datasets domain adaptation should be able to adapt to the evolving statistics. 
Second, starting from associative domain adaptation~\cite{assoc_da} we show that the similar class distribution assumption between domains hurts adaptation.
We therefore reformulate the approach to make the adaptation loss invariant to the inevitable non-stationary changes on the class distribution statistics.
Third, we present two applications of our proposed approach, one on adapting streaming image classification, where the streaming data distribution changes over time, and one on domain adaption for semantic segmentation (see Figure \ref{fig:seg_adapt}), where the source and target datasets have inherently different class statistics.

\section{Related Work}
\label{sec: relwork}


\textbf{Domain Adaptation}   \quad
 A handful of domain adaptation methods revolve around discrepancy-based adaptation \cite{deepCoral, joint_adapt,residual_transfer}, for instance, \cite{MK-MMD} use a multi-kernel maximum mean discrepancy (MMD) minimization approach. Other methods are data reconstruction-based and often use reconstruction with e.g. autoencoders, as an auxiliary task to learn invariant features \cite{recon_bengio, recon_marginalize,recon_nets}.

Another category is adversarial approaches. Adversarial discriminative methods use a classifier to discriminate between domains during training and ensure feature invariance for source and target \cite{dann,ADDA}. Adversarial generative methods use a generative adversarial network (GAN) \cite{GAN} to learn a mapping between source and target images by interleaving the task loss, mapping generator and discriminator loss \cite{pixDA,CyCADA}.  

Domain adaptation for semantic segmentation was recently pioneered by \cite{FCNs_wild} with an adversarial discriminative based approach. Similarly \cite{ROAD} use discriminators for feature invariance, but for different parts of an image grid. \cite{gan_da} use a standard GAN approach to have a generator network learn the mapping while a discriminator network distinguishes between real and fake images. \cite{tri_branch} split the original segmentation network into three output branches where the first two generate pseudo-labels for the third branch. \cite{curr_da} adopt a curriculum learning approach for by solving easy to difficult tasks to achieve adaptation.

\noindent
\textbf{Associative Learning} \quad 
Introduced by \cite{learning_by_assoc}, learning by association was initially applied to semi-supervised learning. \cite{assoc_da} use associations between source and target to close the domain gap for classification achieve competitive results across different domain adaptation benchmarks for classification. The advantage of this method compared to discrepancy-based approaches is that it does not require a choice of kernel and extra hyper-parameters that come with them.

\noindent 
\textbf{Streaming Data Classification} \quad 
Approaches that deal with streaming data are either passive approaches that use a single classifier or an ensemble \cite{ensemble_1,ensemble_2} or active ones where an extra decision is made on whether to update the classifier. Most often classification algorithms such as Decision Trees, Rule-Based and Nearest Neighbor are used, whereas adjustments in neural network architectures to account for streaming have been proposed \cite{survey_}. \cite{sample_filter} use a complex sampling and filtering mechanism for active training and a random forest based classifier. \cite{micro-cluster} use a micro-cluster nearest neighbor which makes use of statistical summaries for data streams. Not many works look into exploiting unsupervised data for improving data stream classifiers. \cite{semisup_stream} use semi-supervised feature learning to adjust k-nearest neighbor weights. To our best knowledge, we are the first to explore this direction for image classification with modern deep architectures.
\section{Method}
\label{sec: method}

\subsection{Associative Domain Adaptation}
\label{sec: assoc}

 \begin{figure*}[t]
\centering
\begin{minipage}{.38\textwidth}
  \centering
  \includegraphics[width=.99\linewidth]{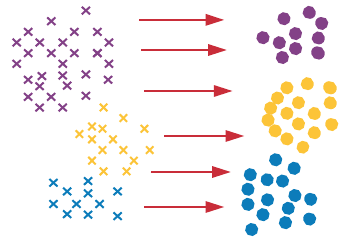}
\end{minipage}%
\hfill
\begin{minipage}{.38\textwidth}
  \centering
  \includegraphics[width=.99\linewidth]{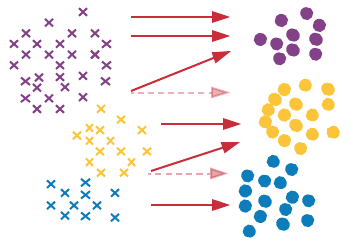}
\end{minipage}
\begin{minipage}{.23\textwidth}
  \centering
  \includegraphics[width=.99\linewidth]{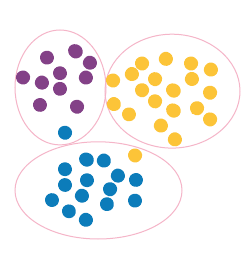}
\end{minipage}
\caption{Associative domain adaptation for unequal class distributions. Crosses represent the source domain and circles represent the target.  Arrows represent source to target probabilities. (a) Uniformly distributed visit loss. (b) Intuition of correcting wrong associations by balancing the visit loss according to class distributions.  (c) Cluster estimates to approximate class distribution in target.}
\label{fig:walker_visit}
\end{figure*}

We start from two datasets, source and target.
The source dataset, $D^{S}=\{x^S_i, y^S_i\}, i=1,..., N^S$, comprises of $N^S$ image samples with embeddings $x^S_i$, annotated by one-hot vectors $y^S_i=[y^S_{ic}], c=1, ..., C$, which equals to 1 if the image $x^S_{ic}$ belongs to class $c$, and 0 otherwise.
The target dataset, $D^{T}=\{x^T_j\}, j=1,...,N^T$, comprises only of image embeddings which belong to the same set of classes, $c=1,...,C$; however, no class annotations are available for retraining or fine-tuning. 
Between the source and target datasets there is a domain shift in the distribution of their respective embeddings,  thus $p(x^S)\neq p(x^T)$.
The goal, therefore, is to adapt a classifier trained on the source dataset to work well for the target. 

Associative domain adaptation \cite{assoc_da} adapts by considering an additional adaptation loss during training on top of the standard task-specific loss, $\mathcal{L} = \mathcal{L}_{task} + \mathcal{L}_{assoc}$.
Specifically, the associative domain adaptation is decomposed into a walker and a visit loss, 
\begin{align}
\mathcal{L}_{assoc} = \mathcal{L}_{walker} + \beta \mathcal{L}_{visit},
\label{eqn:assoc_loss}
\end{align}
where $\beta$ is a weighting coefficient.
Central to the associative domain adaptation is the affinity matrix $A \in \mathbb{R}^{N_S \times N_T}$, which contains elements $a_{ij}$ proportional to how likely the $i$-th embedding in the source domain, $x^S_i$, is to be associated with the $j$-th embedding in the target domain, namely $a_{ij} \propto p(x^T_j|x^S_i)$.
Learning embeddings that yield an affinity matrix that minimizes the loss in Eq.~\eqref{eqn:assoc_loss} is the goal of associative domain adaptation.

The walker and the visit losses have complementary objectives.
The objective of the walker loss is to encourage the source embeddings to lie close after adaptation to source embeddings of the same class.
As no class labels are available in the target dataset, however, this objective is reformulated.
Specifically, after double transition from the source to the target and back to the source, the starting and finishing source class labels should minimize the cross-entropy loss with respect to a normalized equality matrix $E=\{e_{ik}\}$, namely
\begin{align}
\mathcal{L}_{walker} = \sum_{i, k} e_{ik} \log{ \big [ p(x^S_k | x^T_j)\cdot p(x^T_j | x^S_i) \big ]} ,
\label{eqn:walker}
\end{align}
where $x^T_j$ is the closest embedding in the target set and $e_{jk}= \frac{y^S_i \cdot y^S_k}{N_S}$.





The walker loss alone, however, can lead to degenerate solutions, where the transition probabilities are learned to associate source embeddings only with a few relevant yet ``easy'' target embeddings.
To mitigate this, the \textit{visit loss} encourages that all target embeddings are equally visited.
This is achieved by a minimizing cross-entropy objective
\begin{align}
\mathcal{L}_{visit} = \sum_{j} v_j \log{p(x^T_j | x^S_i)},
\label{eqn:visit}
\end{align}
where $v_j=1/N_T$.
\begin{figure*}[t]
  \centering
    \includegraphics[width=0.98\textwidth]{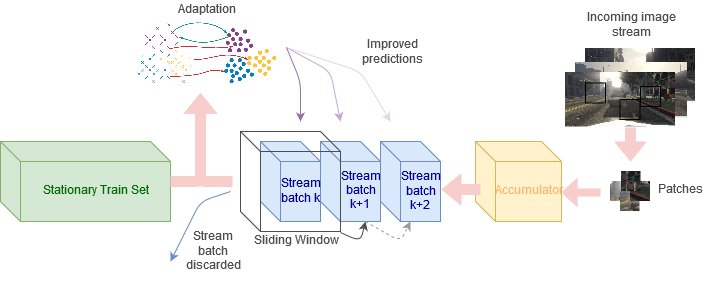}
    \caption{The streaming setup uses a pre-trained model from a stationary supervised set. The model is then sequentially adapted to the incoming stream batches.}
    \label{fig:streaming}
\end{figure*}

\subsection{Associative Domain Adaptation for Unequal Class Distributions}
\label{sec: distrassumption}

Associative domain adaptation implicitly assumes that the source and target distributions are similar on batch level during adaptation. The reason is that for the visit loss to be minimized in Eq.~\eqref{eqn:visit} it is assumed that the ideal target is the average over the size of the target dataset, $v_j=1/N_T$. ~\cite{assoc_da} consider a smaller $\beta$ for the visit loss, if the class distributions between the source and target datasets are unequal. However, this solution implicitly expects access to the adaptation set in order to tune $\beta$. In addition, simply receiving a weaker signal from the visit loss does not exploit the full adaptation capacity and might enforce wrong associations, as we illustrate in Figure \ref{fig:walker_visit} (a). 

As we want target embeddings to be visited by the same-class source embeddings, intuitively they should be visited at a rate proportional to the difference between the source and target class distributions, as shown in Figure \ref{fig:walker_visit} (b). We can formalize the intuition by adding a weighting coefficient in front of $v_j$ and reformulating Eq.~\eqref{eqn:visit} as: 
\begin{align}
\mathcal{L}_{visit} = \sum_{j} \gamma_j v_j \log{p(x^T_j | x^S_i)}, \gamma_j=\frac{p(Y^S =y^T_j)}{p(Y^T=y^T_j)}
\label{eqn:weighted_visit}
\end{align}
namely weighted by the ratio of class probabilities at the source and target for the correct class of the target embedding.
Clearly, we cannot directly compute the ratio $p(Y^S =y^T_j)/p(Y^T=y^T_j)$, as we would need to know the true class of the target embedding $y^T_j$. However, we propose a way to estimate them. 

Although we have no control on the target dataset, we do have control over the source dataset for which the labels are available, thus when constructing the mini-batch based on which we will perform the adaptation, we can first sample the source uniformly such  that all class probabilities are equal in the source dataset, i.e.~$p(Y^S =y^T_j)=const$.
Consequently, from a probabilistic perspective it is not important which particular class the $j$-th target embedding belongs to, alleviating the necessity to make a soft prediction for the class label of the $j$-th embedding.

What remains to compute the weighting coefficient $\gamma_j$ is computing the class probability $p(Y^T =y^T_j)$ for the $j$-th embedding. It is logical to expect that same-class embeddings cluster together for a modern classifier to be able to discriminate between classes. We can retrieve the class cluster around an embedding sample in an unsupervised manner and compute the probability based on cluster size. We rely on unsupervised clustering to estimate class probabilities in the batch. The approximation holds true under the assumption that the clusters are well aligned to the means of the respective, optimal classifiers. In practice, we consider hierarchical agglomerative clustering, which experimentally appears to allow for good alignment between the obtained clusters and works well when clusters have very different sizes. We illustrate the process in Figure \ref{fig:walker_visit} (c).

\subsection{Dynamic Domain Shift in Streaming Data}
\label{sec: assocstr}

Let us consider a pre-collected annotated set $D^{S}=\{x^S_i, y^S_i\}, i=1,..., N^S$ with embeddings $x^S_i$, with one-hot labels $y^S_i$ and an incoming stream of image data that needs to be classified. At every time step $\tau=1,.,K$, the stream is accumulated in a streaming data batch  $D^T_\tau$. Due to the \textit{concept drift}, i.e. distribution shift over time, a classifier $f_0(\theta)$ trained to minimize $\mathcal{L}_{task}(\theta, D^S)$ will perform worse on the streaming batches. Being able to produce accurate predictions as soon as a stream batch comes in is crucial. Changing the models over time aims to account for the concept drift. A second problem to account for in streaming is the size of the incoming data. Usually only a small part of this data can be stored in memory. One way to deal with this is to have a mechanism in place that selects the data to be stored; another way is to be able to use and then discard all the data coming into the stream. 

We simulate a streaming scenario where the stationary training set $D^S$ is pre-collected and first used for off-line training of a predictive model $f_0(\theta)$. Incoming stream data batches $D^{T}_\tau$ are small compared to the stationary set $D^{S}$, but the whole stream cannot be stored in memory, so at a time step $\tau=k$ only a set of $D^T_{k}, D^T_{k+1}, ...D^T_{k+w}$ is available, where w is a storage window size. A classifier $f_{k-1}(\theta)$ trained on the stationary set and adapted to $D^T_1, ... D^T_{k-1}$ sequentially is available. We adapt to $D^T_k$ by minimizing the objective 

\begin{align}
\arg \min_{\substack{\theta}} \mathcal{L}_{task}(f_{k-1}(\theta), y^S) &+ \mathcal{L}_{walker}(\theta, D^S, D^T_k) \nonumber \\ &+ \beta \mathcal{L}_{visit}(\theta, D^S, D^T_k)
\end{align}

The benefits of this approach are twofold. First, adapting to $D^T_k$ improves prediction results on $D^T_k$ itself in an unsupervised manner without extra annotation. Second, the predictions improve for $D^T_{k+1}, D^T_{k+2} ... $ and so on in a cascade fashion, since distribution in incoming sets is more likely to be similar to the previous stream sets nearby than the stationary source, especially if we would use a sliding window over incoming sets. For simplicity we take a window size of 1. An illustration is provided in Figure \ref{fig:streaming}. In our setting, we extract patches from the GTA5 dataset and do patch-wise classification in order to demonstrate the working of our setup with a simpler task. We expect a similar behavior for more complex tasks such as semantic segmentation and object detection.

\subsection{Dynamic Domain Shift in Semantic Segmentation}
\label{sec: assocseg}

Having relaxed the distribution assumption, associative domain adaptation can be applied to tasks where source and target class distributions in a batch are not uniform or uniformity cannot be approximated, such as semantic segmentation. Consider a source dataset $D^{S}=\{x^S_i, y^S_{i, H \times W}\} i=1,..., N^S$, where $H, W$ are image dimensions, is annotated at pixel level. The target images $D^{T}=\{x^T_j\}, j=1,...,N^T$ are available without annotations. Using modern segmentation architectures, we can consider embeddings extracted from a mid-network layer which contains downsampled data. Using a DeepLab-V2 \cite{deeplab_v2} architecture, we extract embedding $x^S_{i'}$ at pixel level in decoder layers before bilinear upsampling which are 8 times downsampled in each spatial dimension. We downsample the label annotations and use  $y^S_{i', U \times V}$, where $U=H/8$ and $V=W/8$ together with downsampled embeddings for adaptation.  

An important consideration when adapting for dense prediction is the choice of affinity matrix $A \in \mathbb{R}^{N_S \times N_T}$ between embeddings, where $a_{ij} \propto p(x^T_j|x^S_i)$. In \cite{assoc_da}, $p(x^T_j|x^S_i)$ is computed as softmax over rows of A, i.e.~
$p(x^T_j|x^S_i) = \exp(a_{ij}) / \sum_{j'} \exp(a_{ij'})$,
where $a_{ij}= x^S_i \cdot x^T_j$ is the dot product between embedding vectors. The unnormalized dot product as an affinity is unbounded and can cause very small probability values for the softmax, which may lead to exploding gradients. We mitigate this by using an affinity measure based on Euclidean distance. 
In addition, we observe that the dimensionality of pixel embeddings for semantic segmentation is crucial for convergence. If too large, the gradients propagated are noisy and adaptation not very effective. However, dimensionality has to be large enough to allow for similar embeddings to group together but still preserve discriminable structures in latent space. For this, we add an \textit{embedding layer} in the decoder where dimensionality can be adjusted for the task.

\section{Experiments and Results}
\label{sec: expres}

We validate the performance of the proposed domain adaptation method under different settings for domain class distribution divergence.  
First, we show the effect of increased class distribution divergence on associative domain adaptation ~\cite{assoc_da} and how we can recover accuracy drops with our formulation. 
Second, we evaluate on a visual stream classification setting, where data and class distributions change over time. 
Third, we further validate the proposed method on domain adaptation for semantic segmentation 
The code, models and datasets will all become available upon publication.

\subsection{Classification under different class distributions between domains}
\label{sec:classificationexp}

We report our results on several image classification adaptation benchmarks. For digit classification we adapt on MNIST \cite{mnist}$\rightarrow$ MNISTM \cite{dann}, SVHN (Street View House Numbers) \cite{svhn} $\rightarrow$ MNIST and Sythetic Digits \cite{dann} to SVHN \cite{svhn}. Next, we adapt for street sign classification from Synthetic Signs dataset \cite{synth_signs} to German Traffic Sign Recognition Benchmark \cite{gtsrb}.  As a last benchmark, CIFAR-10 \cite{cifar10} $\rightarrow$ STL-10 \cite{stl-10} adaptation is performed. Out of the 10 classes present in STL-10 and CIFAR-10, 9 of these overlap so they can be used for domain adaptation. 

We report experiments after changing KL-divergence between the source and target class distributions, to quantify the effect of unequal class distributions for domain adaptation. In Table \ref{tab:distribchange} we report the accuracies over the datasets when class distribution divergence increases for associative domain adaptation, as well as the proposed method.

First, as expected, larger KL-divergence between source and target usually leads to worse accuracy for associative domain adaptation.
Second, the proposed method improves recognition after domain adaptation, especially for larger class distribution divergence, and especially for tasks where the classifiers are not already near maximal adaptation capacity.

To further derive insights, we also include results with an oracle-weighted visit loss that use the target class distributions (theoretical upper bound). 
Although our off-the-shelf agglomerative clustering does not always approximate the batch statistics perfectly, it does come considerably close to the oracle-weighted score and almost always outperforms the unweighted approach. In addition, using oracle test statistics the proposed method often comes close to the recognition accuracies of classifiers trained directly on the target domain indicating that our theoretical reasoning is correct. We conclude that when we expect a dynamical domain shift, where class distributions between the source and target change,our approach is more robust to for domain alignment.

\begin{table}[t]
\caption{Adaptation accuracy as KL-divergence of source to target class distributions in a batch increases. The oracle version uses the true target class probabilities and serves as an upper bound.
}
\label{tab:distribchange}
\begin{center}
\centering
\begin{adjustbox}{max width=\linewidth}
\resizebox{0.98\columnwidth}{!}{%
\begin{tabular}{@{}lllllll@{}}
\toprule
Src -Tgt divergence & Method & \multicolumn{5}{l}{Datasets} \\ \cmidrule(lr){1-1} \cmidrule(lr){2-2} \cmidrule(l){3-7} 
\multirow{3}{*}{} &  & \begin{tabular}[c]{@{}l@{}}MNIST-\\ MNISTM\end{tabular} & \begin{tabular}[c]{@{}l@{}}SVHN-\\ MNIST\end{tabular} & \begin{tabular}[c]{@{}l@{}}Synth Dig.\\ -SVHN\end{tabular} & \begin{tabular}[c]{@{}l@{}}Synth Signs\\ -GTSRB\end{tabular} & \begin{tabular}[c]{@{}l@{}}CIFAR10\\ -STL10\end{tabular} \\ \cmidrule(l){3-7} 
 & Source only & 64.0 & 69.4 & 85.8 & 95.4 & 52.7 \\
 & Target only & 93.6 & 99.5 & 94.2 & 98.1 & 99.8 \\ \midrule
\multirow{3}{*}{KL = 0.05} & Adapted using {\cite{assoc_da}} & 87.6  & 97.0 & 91.9 & 96.2 & 61.3  \\
 & Ours & 88.3  & 97.2 & 92.6 & 96.5 & 61.2  \\
 & Ours with oracle* & 90.0  & 97.2 & 92.8 & 97.5 & 61.5  \\ \midrule
\multirow{3}{*}{KL = 0.2} & Adapted using {\cite{assoc_da}} & 85.2 & 94.3 & 87.6 & 95.9 & 57.6 \\
 & Ours & 87.6 & 96.9 & 89.9 & 95.6 & 58.3 \\
 & Ours with oracle* & 90.1 & 97.8 & 92.8 & 97.3 & 61.2 \\ \midrule
\multirow{3}{*}{KL = 0.4} & Adapted using {\cite{assoc_da}} & 81.7 & 94.2  & 87.1  & 95.5 & 53.4 \\
 & Ours & 83.8 & 94.9 & 88.0 & 95.3 & 56.2 \\
 & Ours with oracle* & 89.8 & 96.6  & 92.6 & 94.1 & 61.4 \\ \bottomrule
\end{tabular}
}
\end{adjustbox}
\end{center}
\end{table}

\subsection{Streaming Data Classification}
\label{sec: strdataexp}

Next, we evaluate the method on a streaming data scenario, where the class distributions are expected to be different between source and target. 
To simulate a streaming data scenario, we note that the popular synthetically generated and finely annotated GTA5 dataset \cite{gta5_paper} is in fact ordered sequentially. Video-like fragments can be observed throughout the dataset, and a shift in distribution over time can also be observed, as shown in Figure \ref{fig:distrib_shift}. We therefore extract patches from GTA5 frame sequences and adapt to a patch-wise classification task, where the label for each patch is equivalent to the dense label for the middle pixel.
We use 65x65 patches cropped from a 256x512 downsampled version of the original GTA5 dataset.

We consider a streaming data scenario where a small set of stationary labeled data is pre-collected and available for training.
For the stationary data, we sample patches from the first 5,000 images in the GTA5 dataset. About 32,000 patches of 65x65 dimensions are sampled.
For the incoming stream we sample patches from bundles of 1,000 images each, collected sequentially.
6,000 patches are sampled from every bundle of images and accumulated in a \textit{streaming batch}.
We experiment with adapting six of these sets following the stationary training set.

Several observations follow from the results in Table \ref{tab:stream-adapt}.
First, there is indeed a dynamical domain shift when considering visual streams instead of static datasets.
When considering the classifiers trained only on the source, there is considerable fluctuation on the recognition accuracy over time.
Note that this is not always harmful, \emph{e.g.}~for streaming batches 5 and 6 accuracy improves, presumably because the shift between target and source is smaller.

Second, the proposed streaming adaptation method yields considerable and constant accuracy improvements over the source-only scores, no matter the source-only recognition accuracy.
Also, the proposed method yields modest but consistent improvements over standard associative domain adaptation \cite{assoc_da}.

Third, as expected, best adaptation is achieved when adapting and testing on the same stream batch (lag = 0).
However, adapting with some lag allows for accurate adaptation as well.
We conclude that for visual streams, where we cannot store the data and we cannot always immediately adapt, dynamical domain adaptation is valuable.

\begin{table}[t]
\caption{Streaming classification accuracy per adaptation round.  Cells marked "-" indicate the batch hasn't yet entered the stream.}
\label{tab:stream-adapt}
\begin{adjustbox}{max width=\linewidth}
\resizebox{0.98\columnwidth}{!}{%
\begin{tabular}{@{}lllllllll@{}}
\toprule
Adaptation Set & Source only  & \multicolumn{6}{l}{Lag from adaptation timestep} & Adapted with \cite{assoc_da}\\  \cmidrule(l){1-1}  \cmidrule(l){2-2}  \cmidrule(l){3-8}  \cmidrule(l){9-9}
 &    & 5  & 4 & 3 & 2 & 1 & 0 &(lag=0)\\ \cmidrule(l){3-8} 
 \rowcolor[HTML]{EFEFEF} 
SB1 & 42.72   & - & - & - & - & - & \textbf{46.72} & 45.58\\
SB2 & 40.30 & - & - & - & - & 44.02 &   \textbf{45.22} & 44.50\\
\rowcolor[HTML]{EFEFEF} 
SB3 & 38.58  & - & -  & -  & 40.88 & 41.48 & 42.02  & \textbf{42.13} \\
SB4 & 37.85  & - &- & 40.88 & 41.52  & 41.98 & \textbf{43.00}  & 42.45\\
\rowcolor[HTML]{EFEFEF} 
SB5 & 42.02  & -  & 45.22  & 45.90 &  45.93  &45.73 &   \textbf{47.51} & 46.13 \\
SB6 & 46.78 & 50.65 & 50.70 & 51.47 & 51.27  &  51.33 & \textbf{52.73} &  51.83   \\ \bottomrule
\end{tabular}
}
\end{adjustbox}
\end{table}

\subsection{Semantic Segmentation}
\label{sec: semanticsegexp}

\begin{table*}[t]
\caption{GTA5 to Cityscapes domain adaptation. The last two rows show results on adapting with the unweighted version of the method and the distribution independent one.}
\label{tab:seg-results}
\begin{center}
\begin{adjustbox}{max width=\linewidth}
\resizebox{0.99\linewidth}{!}{%
\begin{tabular}{@{}llllllllllllllllllllll@{}}
\toprule
& \rotatebox[origin=c]{90}{road} & \rotatebox[origin=c]{90}{sidewalk} & \rotatebox[origin=c]{90}{building} & \rotatebox[origin=c]{90}{wall} & \rotatebox[origin=c]{90}{fence} & \rotatebox[origin=c]{90}{pole} & \rotatebox[origin=c]{90}{light} & \rotatebox[origin=c]{90}{sign} & \rotatebox[origin=c]{90}{vegetation} & \rotatebox[origin=c]{90}{terrain} & \rotatebox[origin=c]{90}{sky} & \rotatebox[origin=c]{90}{person} & \rotatebox[origin=c]{90}{rider} & \rotatebox[origin=c]{90}{car} & \rotatebox[origin=c]{90}{truck} & \rotatebox[origin=c]{90}{bus} & \rotatebox[origin=c]{90}{train} & \rotatebox[origin=c]{90}{motorb.} & \rotatebox[origin=c]{90}{bike} & \cellcolor[HTML]{EFEFEF} mIoU &  \cellcolor[HTML]{EFEFEF} Pixel Acc \\
\midrule

NoAdapt     & 33.8  & 23.2   &  67.5        & 18.2  & \textbf{20.1}    &  18.1   &  15.9          &  \textbf{21.8}     & 66.9    &  18.0   &  72.4      &  33.0     &  6.5   & 25.0      &  \textbf{ 15.8}  &  19.3     &  6.0 & 8.4& \textbf{5.8}    &  \cellcolor[HTML]{EFEFEF} 26.1   & \cellcolor[HTML]{EFEFEF} 72.8 \\
Adapt (no wght.)   & 59.9     & 29.8  & 67.1        &  16.2  &  10.7   &  22.9  &  13.2      & 9.1     &  \textbf{78.0}     &  \textbf{33.4}     &   \textbf{75.7}    &  41.9     & 0.3    & 32.3      &  12.4   &  16.5     &  5.7     &  2.9   &  0.1  &  \cellcolor[HTML]{EFEFEF} 27.8    & \cellcolor[HTML]{EFEFEF} 78.8 \\
Adapt (est. wght.) & \textbf{63.8}    & \textbf{31.3}   &    \textbf{68.4} & \textbf{19.4} & 19.6 & \textbf{23.2}  &  \textbf{17.6} & 11.8  &    62.9     &   22.7   & 61.0    &  \textbf{52.1}    &  \textbf{7.8}   &  \textbf{42.5} &   13.4 & \textbf{22.1}  & \textbf{6.2}   & \textbf{9.1}  & 0.1 & \cellcolor[HTML]{EFEFEF} \textbf{29.2}   &   \cellcolor[HTML]{EFEFEF} \textbf{81.9} \\ 
\bottomrule
\end{tabular}
}
\end{adjustbox}
\end{center}
\end{table*}

Last, we validate the proposed method on the task of domain adaptation for semantic segmentation of urban street scenes. This is an application where source and target class statistics cannot be expected to align, especially on batch level where adaptation happens. 

We adapt on the GTA5 $\rightarrow$ Cityscapes adaptation benchmark, which is important to domain adaptation as adapting from synthetic to real data provides potential for exploiting very easily rendered synthetic sets. GTA5 contains 24,966 images with resolution 1914$\times$1052, of which 12,500 are used for training and around 6,800 for validation. Cityscapes contains 5,000 pixel-level annotated images of 2048$\times$1024 resolution, of which 2,975 images for the training set and 500 images for validation are available. We run our experiments with images from both datasets downsampled to 512$\times$256 size. 

As a base segmentation network we use DeepLab-V2 \cite{deeplab_v2} with a ResNet-50 \cite{resnet-50} backbone. We extend the original DeepLab-V2 architecture with a $D$-dimensional embedding layer that can be adjusted for experiment purposes and report results with $D=64$. The embedding layer is placed before the bilinear upsampling part of the decoder, yielding embeddings that are 8 times downsampled in each spatial dimension. In this way we can not only adapt to more compressed information on pixel level embeddings, but also fit embedding metrics in reasonable memory even for large datasets. 

We use $\beta=0.5$ for the visit loss, adjusted for the magnitude of the loss values.
We use the respective training sets of GTA5 and Cityscapes as the domains for training, test on the Cityscapes \textit{val} set, and report the results in Table \ref{tab:seg-results}.

First, we train for 30K iterations on source only for GTA5, using pre-trained ImageNet \cite{imagenet} weights for the ResNet-50 encoder part of the network. We observe that the proposed distribution independent approach consistently improves standard associative domain adaptation, both in terms of mIoU and pixel accuracy.

Further, the proposed method improves standard domain adaptation on 15 out of the 19 categories.
Standard associative domain adaptation is still better for large classes with near constant class frequency (e.g.~\emph{vegetation}, \emph{terrain}, \emph{sky}), since adaptation over these would overrule smaller classes in a batch. Interestingly, the proposed method seem to improve significantly ($6-10\%$) over mid-size classes, such as \textit{car}, \textit{bus} and \textit{person}, where indeed we expect larger class frequency fluctuations. 
We conclude that our approach is promising for domain adaptation of complex dense prediction tasks such as semantic segmentation, and potentially, integrating with the streaming techniques above, to video semantic segmentation. 



\section{Conclusion}
\label{sec: conclusion}

We have presented a robust and distribution independent associative learning method for domain adaptation. Our formulation accounts for realistic scenarios where source and target data distribution in a batch cannot be approximated to be equal. A novel setup for dynamic domain adaptation that adapts over unlabeled data in order to improve classifier prediction over time for streaming data has been proposed. We have shown that we can exploit unsupervised data to achieve improvements over several streaming batches without additionally annotated samples. Using our associative domain adaptation formulation and architecture considerations we achieve competitive results for semantic segmentation.

Having considered a dynamic time-shifting distribution setup and shown dense prediction adaptation results, we lay the grounds for a framework that can potentially work well with dense prediction tasks for streaming video data such as video segmentation.

{\small
\bibliographystyle{ieee}
\bibliography{bibliography/bibliography.bib}
}

\end{document}